\documentclass[11pt]{article}

\usepackage[utf8]{inputenc}
\usepackage[T1]{fontenc}
\usepackage[margin=1in]{geometry}
\usepackage[numbers,sort&compress]{natbib}
\usepackage{hyperref}
\usepackage{url}
\usepackage{booktabs}
\usepackage{amsmath}
\usepackage{amsfonts}
\usepackage{graphicx}
\usepackage{subcaption}
\usepackage{multirow}
\usepackage{float}
\usepackage{microtype}
\usepackage{cleveref}
\usepackage{xcolor}
\usepackage{nicefrac}
\usepackage{tabularx}
\usepackage{array}
\usepackage{enumitem}

\microtypesetup{expansion=false}

\hypersetup{
  colorlinks=true,
  linkcolor=blue,
  citecolor=blue,
  urlcolor=blue
}

\title{\textbf{Fine-Grained Action Segmentation for Renorrhaphy in Robot-Assisted Partial Nephrectomy}}
\author{\begin{tabular}{c}
\Large Jiaheng Dai$^{1,*}$, Huanrong Liu$^{2,*}$, Tailai Zhou$^{3,*}$, Tongyu Jia$^{3,*}$ \\
\Large Qin Liu$^2$, Yutong Ban$^4$, Zeju Li$^1$ \\
\Large Yu Gao$^3$, Xin Ma$^3$, Qingbiao Li$^2$ \\[0.6em]
\normalsize $^1$Fudan University, Shanghai, China \\
\normalsize $^2$University of Macau, Macau, China \\
\normalsize $^3$The Chinese PLA General Hospital, Beijing, China \\
\normalsize $^4$Shanghai Jiao Tong University, Shanghai, China
\end{tabular}}
\date{}

\begin{document}

\maketitle
\begingroup
\renewcommand{\thefootnote}{*}
\footnotetext{Equal contribution.}
\endgroup

\begin{abstract}
Fine-grained action segmentation during renorrhaphy in robot-assisted partial nephrectomy requires frame-level recognition of visually similar suturing gestures with variable duration and substantial class imbalance. The SIA-RAPN benchmark defines this problem on 50 clinical videos acquired with the da Vinci Xi system and annotated with 12 frame-level labels. The benchmark compares four temporal models built on I3D features: MS-TCN++, AsFormer, TUT, and DiffAct. Evaluation uses balanced accuracy, edit score, segmental F1 at overlap thresholds of 10, 25, and 50, frame-wise accuracy, and frame-wise mean average precision. In addition to the primary evaluation across five released split configurations on SIA-RAPN, the benchmark reports cross-domain results on a separate single-port RAPN dataset. Across the strongest reported values over those five runs on the primary dataset, DiffAct achieves the highest F1, frame-wise accuracy, edit score, and frame mAP, while MS-TCN++ attains the highest balanced accuracy.
\end{abstract}

\section{Introduction}

Robot-assisted partial nephrectomy (RAPN) is a common organ-preserving procedure in urologic surgery. Renorrhaphy, the reconstructive stage that closes the renal resection bed, is technically demanding because it must restore tissue integrity while maintaining hemostasis and limiting warm ischemia time \citep{falagario2020robotic,van2023suturing}. Automatic analysis of this stage is relevant to surgical training, retrospective workflow assessment, and the development of video-based assistance systems.

Fine-grained surgical activity analysis has been studied in both laboratory and clinical robotic settings, including benchmark datasets for gesture recognition and action recognition from minimally invasive surgery \citep{gao2014jhu,itzkovich2021generalization,psychogyios2023sar}. Renorrhaphy remains difficult for temporal segmentation because adjacent gestures differ by subtle tool--tissue interactions, rare events are clinically meaningful, and visual conditions vary across patients and surgeons.

SIA-RAPN formulates this problem as frame-level action segmentation on untrimmed renorrhaphy videos. Each frame is assigned one of 12 labels, and evaluation requires both correct semantic labeling and accurate temporal localization. The task therefore couples short-range motion discrimination with longer-range reasoning over action order, duration, and boundary ambiguity.

The benchmark makes three contributions:
\begin{itemize}[leftmargin=1.5em]
    \item It defines a clinically grounded action-segmentation task for renorrhaphy using 50 curated clinical RAPN videos annotated with 12 frame-level labels.
    \item It establishes a benchmark protocol with frame-level and segment-level metrics designed to capture class imbalance, temporal boundary quality, and sequence consistency.
    \item It reports baseline results for multiple temporal modeling families, including convolutional, Transformer-based, and diffusion-based approaches, under both in-domain evaluation across five released split configurations and cross-domain evaluation on a separate single-port RAPN cohort.
\end{itemize}

\section{Task Definition and Dataset}

\subsection{Primary cohort and task definition}

SIA-RAPN is derived from robot-assisted partial nephrectomy procedures collected at the Chinese PLA General Hospital. The broader study enrolled 207 patients, and 50 videos were retained for benchmark construction after clinical and technical quality control. The recordings were acquired with the da Vinci Xi system at 1080i and 60 fps and preserve realistic variation in visibility, bleeding, camera viewpoint, and resection-bed appearance. The benchmark uses one stereo stream per case, prioritizing the left-eye view and substituting the right-eye view when the left-eye stream is severely contaminated. All participants provided written informed consent, and the study was approved by the Ethics Committee of the Chinese PLA General Hospital (approval number S2024-225-01). The task addresses \textbf{action segmentation} during renorrhaphy. Given a long untrimmed video segment, the model must assign a label to each frame and recover the temporal boundaries of the surgical gestures. The resulting prediction should therefore be both \emph{semantically correct} and \emph{temporally consistent}. Compared with generic activity-recognition datasets, the task is challenging because adjacent suturing gestures can differ by subtle tool--tissue interactions, clinically meaningful events are often rare, and gesture duration varies substantially across the procedure.

\subsection{Annotation, label release, and supplementary cohort}

Frame-level action intervals were annotated from the surgical videos and exported as structured label sequences for model training and evaluation. Annotation was performed by trained urologists and subsequently reviewed by senior clinicians. The benchmark uses 12 frame-level labels in total. This paper therefore reports the task setup, evaluation protocol, and benchmark results without enumerating the full label taxonomy. In addition to the primary SIA-RAPN benchmark, we report results on a separate single-port RAPN dataset. This cohort is treated as a cross-domain evaluation set rather than as an additional fold of the main benchmark, and it is used to probe robustness under a domain shift in platform characteristics and visual appearance. Related work on anatomy segmentation, articulation prediction, surgical action recognition, and motion estimation in robot-assisted partial nephrectomy (ID: AM26-2507/A0335) will be presented at EAU26 (\url{https://eaucongress.uroweb.org/}). Two properties are especially relevant for modeling: class imbalance across labels and substantial variability in action duration. Together, they motivate the use of balanced metrics, class-reweighted objectives, and temporal models that can operate over both short and long contexts.

\section{Benchmark Models}

The benchmark combines a frame-level video encoder with temporal models that map those features to frame-wise gesture labels. All temporal baselines reported here use I3D features.

\subsection{Spatial feature encoder}

\paragraph{I3D.}
Inflated 3D ConvNet (I3D) inflates 2D convolutional filters into 3D to encode appearance and short-range motion jointly \citep{carreira2017quo}. It is relevant here as a conventional video backbone that provides a common feature representation for temporal decoders.

\subsection{Temporal decoders}

\paragraph{MS-TCN++.}
MS-TCN++ refines temporal predictions over multiple stages with dilated temporal convolutions and iterative refinement \citep{liu2020ms}. It provides a strong convolutional baseline for long surgical sequences with noisy frame-level features.

\paragraph{AsFormer.}
AsFormer is a Transformer-based temporal action segmentation model with adaptive attention spans for long-range temporal reasoning \citep{yi2021asformer}. It is relevant to this task because similar suturing gestures often require broader temporal context to disambiguate.

\paragraph{Temporal U-Transformer (TUT).}
Temporal U-Transformer (TUT) is a U-shaped Transformer architecture proposed for action segmentation with multi-scale temporal modeling and a boundary-aware training formulation \citep{du2023we}. It is relevant as a pure Transformer alternative that emphasizes multi-scale temporal structure without temporal convolutions.

\paragraph{DiffAct.}
DiffAct formulates action segmentation as iterative denoising conditioned on video features \citep{liu2024diffact++}. It is relevant because repeated refinement can help with boundary ambiguity and long-range action dependencies in suturing sequences.

\subsection{Benchmark configuration}

The reported benchmark compares four temporal models on a common I3D feature representation: MS-TCN++, AsFormer, TUT, and DiffAct. This design isolates differences in temporal modeling while keeping the frame-level encoder fixed.

\section{Experimental Protocol}
\label{sec:exp_protocol}

\subsection{Data split and evaluation setup}

The released processed dataset contains 54 labeled base-case files on disk, of which 50 are used in the benchmark protocol. For each released split configuration, 41 cases are used for training, 5 for validation, and a fixed 4-case held-out set is used for testing. The same held-out test set is used across splits 1 to 5. The per-split tables in the results section therefore report five independently trained runs, while the final summary table retains the strongest reported value for each metric across those five runs.

We additionally report results on a separate single-port RAPN dataset. This dataset is treated as a cross-domain evaluation set rather than as part of the primary split protocol. The corresponding tables are reported separately to distinguish in-domain performance on SIA-RAPN from cross-domain performance on the single-port cohort.

\subsection{Data augmentation}

To improve robustness to visual variability, the training data were augmented with rotation, brightness and contrast perturbation, blur, and optional cutout. The corresponding labels and feature sequences were kept temporally aligned during preprocessing before temporal-model training.

\subsection{Metrics}

Let $T$ denote the number of frames in a video, $C$ the number of evaluated labels, $y_t \in \{1,\dots,C\}$ the ground-truth label at frame $t$, and $\hat{y}_t$ the predicted label. The benchmark reports the following frame-level and segment-level metrics.

\paragraph{Frame-wise accuracy.}
\begin{equation}
\mathrm{Acc} = \frac{1}{T} \sum_{t=1}^{T} \mathbf{1}\!\left[\hat{y}_t = y_t\right].
\end{equation}
This metric measures the fraction of correctly labeled frames.

\paragraph{Balanced accuracy.}
Balanced accuracy is the macro-average of per-class recall \citep{brodersen2010balanced}:
\begin{equation}
\mathrm{BalAcc} = \frac{1}{C} \sum_{c=1}^{C} \frac{\mathrm{TP}_c}{\mathrm{TP}_c + \mathrm{FN}_c}.
\end{equation}
It reduces the dominance of frequent classes and is therefore important for imbalanced surgical action data.

\paragraph{Edit score.}
Following \citet{lea2017temporal}, let $\mathcal{C}(\cdot)$ collapse consecutive repeated labels into a segment sequence and let $\mathrm{Lev}(\cdot,\cdot)$ denote Levenshtein distance. The normalized edit score is
\begin{equation}
\mathrm{Edit} = 1 - \frac{\mathrm{Lev}\!\left(\mathcal{C}(\hat{y}_{1:T}), \mathcal{C}(y_{1:T})\right)}{\max\!\left(\left|\mathcal{C}(\hat{y}_{1:T})\right|, \left|\mathcal{C}(y_{1:T})\right|\right)}.
\end{equation}
This metric emphasizes sequence consistency and penalizes over-segmentation.

\paragraph{Segmental F1 at IoU thresholds 0.10, 0.25, and 0.50.}
Following the segment-based evaluation protocol commonly used in temporal action segmentation \citep{lea2017temporal}, the benchmark reports three separate segmental F1 scores, denoted F1@10, F1@25, and F1@50. For a given overlap threshold $\tau \in \{0.10, 0.25, 0.50\}$, the evaluation first collapses frame-wise labels into contiguous temporal segments. A predicted segment contributes to $\mathrm{TP}_{\tau}$ when its best same-class intersection-over-union reaches $\tau$; otherwise it contributes to $\mathrm{FP}_{\tau}$. Ground-truth segments whose best same-class overlap stays below $\tau$ contribute to $\mathrm{FN}_{\tau}$. Precision, recall, and F1 are
\begin{equation}
\mathrm{Precision}_{\tau} = \frac{\mathrm{TP}_{\tau}}{\mathrm{TP}_{\tau} + \mathrm{FP}_{\tau}}, \qquad
\mathrm{Recall}_{\tau} = \frac{\mathrm{TP}_{\tau}}{\mathrm{TP}_{\tau} + \mathrm{FN}_{\tau}},
\end{equation}
\begin{equation}
\mathrm{F1@}\tau = \frac{2 \, \mathrm{Precision}_{\tau} \, \mathrm{Recall}_{\tau}}{\mathrm{Precision}_{\tau} + \mathrm{Recall}_{\tau}}.
\end{equation}
These scores quantify temporal boundary quality at increasingly strict overlap thresholds.

\paragraph{Frame-wise mean average precision.}
Following the standard average-precision formulation based on precision--recall evaluation \citep{davis2006relationship}, frame mAP is computed from one-hot frame labels rather than from continuous confidence scores. Let $y_t^c = \mathbf{1}[y_t = c]$ and $\hat{y}_t^c = \mathbf{1}[\hat{y}_t = c]$. For each class $c$, average precision $\mathrm{AP}_c$ is computed in a one-vs-rest manner from the pairs $\{(y_t^c,\hat{y}_t^c)\}_{t=1}^{T}$, and frame-wise mean average precision is then
\begin{equation}
\mathrm{mAP} = \frac{1}{C} \sum_{c=1}^{C} \mathrm{AP}_c.
\end{equation}
This metric complements accuracy-based measures by summarizing per-class frame retrieval quality under the released evaluation implementation.

\subsection{Model-specific implementation and engineering details}

All temporal baselines operate on a shared I3D RGB feature representation with 1024-dimensional descriptors. Using a common visual backbone allows the comparison to focus on temporal modeling rather than on differences in frame-level encoding.

Beyond matching the nominal model families from prior work, the released benchmark also reflects a common layer of engineering adaptation for long surgical videos. All models use the same processed features, label mapping, and released train/validation partitions, and all checkpoints are selected according to validation-set performance. The resulting comparison therefore captures both architectural differences and the practical optimization choices needed to train reliably under class imbalance, boundary ambiguity, and limited sample size.

\paragraph{I3D feature pipeline.}
The common feature extractor uses an ImageNet-pretrained RGB I3D backbone \citep{carreira2017quo} as a fixed feature extractor without further fine-tuning during temporal-model training. This produces the same 1024-dimensional representation for all four temporal baselines and keeps the backbone and label mapping fixed across model families.

\paragraph{MS-TCN++.}
MS-TCN++ serves as the convolutional baseline and is configured to emphasize stable temporal refinement on long videos. The benchmark uses a deep multi-stage design with 13 prediction-generation layers, 12 refinement layers per stage, and 3 refinement stages, trained for 200 epochs with Adam and an initial learning rate of $3 \times 10^{-4}$. In practice, the combination of class-reweighted classification and temporal smoothing helps reduce fragmented frame-wise predictions while remaining robust to the uneven gesture distribution of the suturing task.

\paragraph{AsFormer.}
AsFormer is implemented as a Transformer baseline for stronger long-range temporal reasoning, but its training is regularized to remain stable on the relatively small surgical dataset. The adopted configuration uses 10 layers, 64 feature maps, a 1024-dimensional input stream, channel masking at rate 0.3, and one encoder followed by three decoder stages. It is trained for 150 epochs with Adam and an initial learning rate of $1 \times 10^{-4}$. In addition, augmented training samples are included so that the model can benefit from greater appearance diversity without changing the underlying backbone features.

\paragraph{TUT.}
TUT is used as a multi-scale Transformer alternative and is adapted to the surgical setting through segmented temporal training. The released configuration uses window size 31 and segment length 200 together with five prediction-generation layers, five refinement layers, and three refinement stages, with hidden sizes 64 and 32 for prediction and refinement. Training runs for 200 epochs with batch size 1, Adam, and an initial learning rate of $3 \times 10^{-4}$. In the released surgery setup, the effective objective is cross-entropy with temporal smoothing ($\gamma = 0.15$), which provides a more stable optimization path on long sequences while preserving the model's coarse-to-fine temporal structure.

\paragraph{DiffAct.}
DiffAct is the most elaborate temporal decoder in the benchmark and couples denoising-based refinement with several practical stabilizers. The released surgery runs use split-specific settings with sample rate 2, temporal augmentation, an encoder with 14 layers and 128 feature maps, a decoder with 10 layers and 64 feature maps, diffusion length 1000 with 25 sampling steps, batch size 2, and learning rate $5 \times 10^{-4}$. Class weighting and temporal post-processing are retained in the pipeline because they improve robustness to imbalance and help produce cleaner segment boundaries after iterative refinement.

\section{Results}

\subsection{Primary evaluation across five released splits on SIA-RAPN}

\Cref{tab:best_model_performance} preserves the strongest reported value for each metric across the five primary-dataset splits. DiffAct attains the highest F1@10, F1@25, F1@50, frame-wise accuracy, edit score, and frame mAP, whereas MS-TCN++ attains the highest balanced accuracy. AsFormer yields the second-best summary values on F1@10, F1@25, and edit score, while TUT remains competitive with AsFormer on F1@50.

Across the detailed per-split results, DiffAct remains the strongest model on the primary dataset, while MS-TCN++ is more competitive on balanced accuracy. This contrast is consistent with the benchmark design: DiffAct benefits from iterative temporal refinement, whereas MS-TCN++ remains a strong class-balanced convolutional baseline.

\subsection{Single-port cross-domain evaluation}

The single-port results are reported separately because they represent a different surgical domain rather than additional SIA-RAPN split configurations. In this cross-domain setting, DiffAct remains strongest on F1@25, F1@50, frame-wise accuracy, balanced accuracy, edit score, and frame mAP, while AsFormer attains the highest F1@10 in the reported single-port runs.

\subsection{Detailed Benchmark Tables}

\begin{table}[H]
\centering
\caption{Summary comparison using the strongest reported value for each metric across the five splits.}
\label{tab:best_model_performance}
\begin{tabular}{lccccccc}
\toprule
\textbf{Model} & \textbf{F1@10} & \textbf{F1@25} & \textbf{F1@50} & \textbf{Acc} & \textbf{Bal-Acc} & \textbf{Edit} & \textbf{Frame mAP} \\
\midrule
MS-TCN++ & 76.54 & 71.96 & 58.02 & 68.83 & \textbf{67.54} & 74.93 & 48.79 \\
AsFormer & 79.60 & 73.78 & 57.87 & 64.55 & 58.73 & 78.00 & 40.08 \\
TUT & 76.04 & 71.03 & 57.62 & 61.86 & 54.57 & 76.28 & 37.61 \\
DiffAct & \textbf{83.14} & \textbf{80.23} & \textbf{66.86} & \textbf{72.12} & 67.16 & \textbf{82.30} & \textbf{51.28} \\
\bottomrule
\end{tabular}
\end{table}

\begin{table}[H]
\centering
\small
\caption{AsFormer per-split performance on the primary SIA-RAPN benchmark.}
\label{tab:asformer_split_performance}
\begin{tabular}{lccccc}
\toprule
\textbf{Split} & \textbf{F1@\{10,25,50\}} & \textbf{Acc} & \textbf{Bal-Acc} & \textbf{Edit} & \textbf{Frame mAP} \\
\midrule
Split 1 & 0.7845 / 0.7378 / 0.5635 & 0.6348 & 0.5752 & 0.7568 & 0.3936 \\
Split 2 & 0.7804 / 0.7147 / 0.5751 & 0.6302 & 0.5664 & 0.7800 & 0.3927 \\
Split 3 & 0.7131 / 0.6346 / 0.4477 & 0.5498 & 0.4953 & 0.7271 & 0.3079 \\
Split 4 & 0.7653 / 0.6937 / 0.5279 & 0.6304 & 0.5873 & 0.7461 & 0.4008 \\
Split 5 & 0.7960 / 0.7309 / 0.5787 & 0.6455 & 0.5745 & 0.7872 & 0.3968 \\
\bottomrule
\end{tabular}
\end{table}

\begin{table}[H]
\centering
\small
\caption{TUT per-split performance on the primary SIA-RAPN benchmark.}
\label{tab:tut_split_performance}
\begin{tabular}{lccccc}
\toprule
\textbf{Split} & \textbf{F1@\{10,25,50\}} & \textbf{Acc} & \textbf{Bal-Acc} & \textbf{Edit} & \textbf{Frame mAP} \\
\midrule
Split 1 & 0.7604 / 0.7103 / 0.5762 & 0.6186 & 0.5457 & 0.7628 & 0.3761 \\
Split 2 & 0.7498 / 0.6885 / 0.5415 & 0.6147 & 0.5447 & 0.7242 & 0.3677 \\
Split 3 & 0.7295 / 0.6749 / 0.5076 & 0.5975 & 0.5167 & 0.7364 & 0.3398 \\
Split 4 & 0.7361 / 0.6729 / 0.5279 & 0.5866 & 0.5119 & 0.7542 & 0.3471 \\
Split 5 & 0.7421 / 0.7022 / 0.5567 & 0.5992 & 0.5203 & 0.7421 & 0.3621 \\
\bottomrule
\end{tabular}
\end{table}

\begin{table}[H]
\centering
\small
\caption{MS-TCN++ per-split performance on the primary SIA-RAPN benchmark.}
\label{tab:mstcnpp_split_performance}
\begin{tabular}{lccccc}
\toprule
\textbf{Split} & \textbf{F1@\{10,25,50\}} & \textbf{Acc} & \textbf{Bal-Acc} & \textbf{Edit} & \textbf{Frame mAP} \\
\midrule
Split 1 & 0.7491 / 0.6961 / 0.5318 & 0.6446 & 0.6120 & 0.7357 & 0.4203 \\
Split 2 & 0.7542 / 0.7149 / 0.5630 & 0.6651 & 0.5994 & 0.7426 & 0.4300 \\
Split 3 & 0.7609 / 0.7144 / 0.5637 & 0.6836 & 0.6534 & 0.7477 & 0.4768 \\
Split 4 & 0.7259 / 0.6879 / 0.5534 & 0.6499 & 0.5866 & 0.7200 & 0.4148 \\
Split 5 & 0.7654 / 0.7196 / 0.5802 & 0.6883 & 0.6754 & 0.7493 & 0.4879 \\
\bottomrule
\end{tabular}
\end{table}

\begin{table}[H]
\centering
\small
\caption{DiffAct per-split performance on the primary SIA-RAPN benchmark.}
\label{tab:diffact_split_performance}
\begin{tabular}{lccccc}
\toprule
\textbf{Split} & \textbf{F1@\{10,25,50\}} & \textbf{Acc} & \textbf{Bal-Acc} & \textbf{Edit} & \textbf{Frame mAP} \\
\midrule
Split 1 & 0.8274 / 0.8023 / 0.6654 & 0.7166 & 0.6541 & 0.8158 & 0.5017 \\
Split 2 & 0.8012 / 0.7600 / 0.6268 & 0.6679 & 0.5982 & 0.8021 & 0.4291 \\
Split 3 & 0.8207 / 0.7836 / 0.6686 & 0.7103 & 0.6573 & 0.8230 & 0.4942 \\
Split 4 & 0.8314 / 0.7926 / 0.6589 & 0.7059 & 0.6563 & 0.8132 & 0.4861 \\
Split 5 & 0.8291 / 0.7918 / 0.6536 & 0.7212 & 0.6716 & 0.8205 & 0.5128 \\
\bottomrule
\end{tabular}
\end{table}

\begin{table}[H]
\centering
\small
\caption{DiffAct per-split performance on the single-port cross-domain dataset.}
\label{tab:singleport_diffact_performance}
\begin{tabular}{lccccc}
\toprule
\textbf{Split} & \textbf{F1@\{10,25,50\}} & \textbf{Acc} & \textbf{Bal-Acc} & \textbf{Edit} & \textbf{Frame mAP} \\
\midrule
Split 1 & 0.7414 / 0.6729 / 0.4361 & 0.5985 & 0.5157 & 0.7547 & 0.2947 \\
Split 2 & 0.7099 / 0.6049 / 0.4444 & 0.5724 & 0.5040 & 0.7368 & 0.3066 \\
Split 3 & 0.7203 / 0.6302 / 0.4759 & 0.6009 & 0.5369 & 0.6984 & 0.3129 \\
Split 4 & 0.7143 / 0.6211 / 0.4596 & 0.5731 & 0.5855 & 0.6749 & 0.3179 \\
Split 5 & 0.7055 / 0.6196 / 0.3926 & 0.5894 & 0.5270 & 0.7523 & 0.3084 \\
\bottomrule
\end{tabular}
\end{table}

\begin{table}[H]
\centering
\small
\caption{AsFormer per-split performance on the single-port cross-domain dataset.}
\label{tab:singleport_asformer_performance}
\begin{tabular}{lccccc}
\toprule
\textbf{Split} & \textbf{F1@\{10,25,50\}} & \textbf{Acc} & \textbf{Bal-Acc} & \textbf{Edit} & \textbf{Frame mAP} \\
\midrule
Split 1 & 0.7439 / 0.6201 / 0.4012 & 0.5489 & 0.5479 & 0.7514 & 0.2835 \\
Split 2 & 0.6748 / 0.5636 / 0.3273 & 0.4980 & 0.4031 & 0.7241 & 0.2315 \\
Split 3 & 0.6647 / 0.5612 / 0.2866 & 0.4400 & 0.3782 & 0.6592 & 0.1962 \\
Split 4 & 0.6890 / 0.5775 / 0.3222 & 0.5060 & 0.5425 & 0.6994 & 0.2630 \\
Split 5 & 0.7089 / 0.5849 / 0.3270 & 0.4876 & 0.4548 & 0.7531 & 0.2229 \\
\bottomrule
\end{tabular}
\end{table}

\begin{table}[H]
\centering
\small
\caption{TUT per-split performance on the single-port cross-domain dataset.}
\label{tab:singleport_tut_performance}
\begin{tabular}{lccccc}
\toprule
\textbf{Split} & \textbf{F1@\{10,25,50\}} & \textbf{Acc} & \textbf{Bal-Acc} & \textbf{Edit} & \textbf{Frame mAP} \\
\midrule
Split 1 & 0.6609 / 0.5698 / 0.3605 & 0.5383 & 0.4811 & 0.6170 & 0.2763 \\
Split 2 & 0.6952 / 0.6132 / 0.4126 & 0.5249 & 0.4767 & 0.6425 & 0.2816 \\
Split 3 & 0.6301 / 0.5533 / 0.3919 & 0.4820 & 0.4051 & 0.6597 & 0.2183 \\
Split 4 & 0.6762 / 0.5886 / 0.3438 & 0.5508 & 0.5221 & 0.6684 & 0.2781 \\
Split 5 & 0.6592 / 0.5690 / 0.3662 & 0.5407 & 0.4791 & 0.6683 & 0.2738 \\
\bottomrule
\end{tabular}
\end{table}

\begin{table}[H]
\centering
\small
\caption{MS-TCN++ per-split performance on the single-port cross-domain dataset.}
\label{tab:singleport_mstcnpp_performance}
\begin{tabular}{lccccc}
\toprule
\textbf{Split} & \textbf{F1@\{10,25,50\}} & \textbf{Acc} & \textbf{Bal-Acc} & \textbf{Edit} & \textbf{Frame mAP} \\
\midrule
Split 1 & 0.5277 / 0.4286 / 0.2468 & 0.4330 & 0.4221 & 0.5577 & 0.2078 \\
Split 2 & 0.5696 / 0.4401 / 0.2330 & 0.4710 & 0.4470 & 0.5192 & 0.2264 \\
Split 3 & 0.5263 / 0.4637 / 0.2561 & 0.4754 & 0.4401 & 0.5449 & 0.2326 \\
Split 4 & 0.4986 / 0.4000 / 0.2197 & 0.4721 & 0.4219 & 0.5053 & 0.2283 \\
Split 5 & 0.4354 / 0.3618 / 0.2041 & 0.3864 & 0.3474 & 0.4231 & 0.1726 \\
\bottomrule
\end{tabular}
\end{table}

\section{Discussion}

Several limitations should be kept in view when interpreting the benchmark. The dataset contains 50 curated cases from a single collection site, which limits immediate claims about generalization across institutions, surgeons, and acquisition conditions. Although results are reported across five splits, the summary comparison table aggregates the strongest reported value for each metric rather than representing a single fold. Broader validation on additional cohorts would strengthen external validity.

\section{Conclusion}

Fine-grained action segmentation for renorrhaphy in RAPN combines subtle gesture boundaries, class imbalance, and substantial clinical variability. The benchmark defines a 12-label frame-level recognition problem on 50 clinical videos and reports baseline results for four temporal models built on I3D features. Across the strongest reported values over five released split configurations, DiffAct attains the highest F1, frame-wise accuracy, edit score, and frame mAP, while MS-TCN++ attains the highest balanced accuracy. These results provide a quantitative reference for future work on temporally precise surgical video understanding in renorrhaphy.

\bibliographystyle{unsrtnat}
\bibliography{references}

@article{falagario2020robotic,
  title={Robotic-assisted surgery for the treatment of urologic cancers: recent advances},
  author={Falagario, Ugo and Veccia, Alessandro and Weprin, Samuel and Albuquerque, Emanuel V and Nahas, William C and Carrieri, Giuseppe and Pansadoro, Vito and Hampton, Lance J and Porpiglia, Francesco and Autorino, Riccardo},
  journal={Expert Review of Medical Devices},
  volume={17},
  number={6},
  pages={579--590},
  year={2020},
  publisher={Taylor \& Francis}
}

@incollection{van2023suturing,
  title={Suturing Techniques in Robot-Asssisted Partial Nephrectomy (RAPN)},
  author={Van Puyvelde, Hannah and De Groote, Ruben},
  booktitle={Robotic Surgery for Renal Cancer},
  pages={1--5},
  year={2023},
  publisher={Springer}
}

@article{psychogyios2023sar,
  title={Sar-rarp50: Segmentation of surgical instrumentation and action recognition on robot-assisted radical prostatectomy challenge},
  author={Psychogyios, Dimitrios and Colleoni, Emanuele and Van Amsterdam, Beatrice and Li, Chih-Yang and Huang, Shu-Yu and Li, Yuchong and Jia, Fucang and Zou, Baosheng and Wang, Guotai and Liu, Yang and others},
  journal={arXiv preprint arXiv:2401.00496},
  year={2023}
}

@article{itzkovich2021generalization,
  title={Generalization of deep learning gesture classification in robotic-assisted surgical data: From dry lab to clinical-like data},
  author={Itzkovich, Danit and Sharon, Yarden and Jarc, Anthony and Refaely, Yael and Nisky, Ilana},
  journal={IEEE Journal of Biomedical and Health Informatics},
  volume={26},
  number={3},
  pages={1329--1340},
  year={2021},
  publisher={IEEE}
}

@inproceedings{gao2014jhu,
  title={Jhu-isi gesture and skill assessment working set (jigsaws): A surgical activity dataset for human motion modeling},
  author={Gao, Yixin and Vedula, S Swaroop and Reiley, Carol E and Ahmidi, Narges and Varadarajan, Balakrishnan and Lin, Henry C and Tao, Lingling and Zappella, Luca and B{\'e}jar, Benjam{\i}n and Yuh, David D and others},
  booktitle={MICCAI workshop: M2cai},
  pages={3},
  year={2014}
}

@article{yi2021asformer,
  title={Asformer: Transformer for action segmentation},
  author={Yi, Fangqiu and Wen, Hongyu and Jiang, Tingting},
  journal={arXiv preprint arXiv:2110.08568},
  year={2021}
}

@inproceedings{carreira2017quo,
  title={Quo vadis, action recognition? a new model and the kinetics dataset},
  author={Carreira, Joao and Zisserman, Andrew},
  booktitle={proceedings of the IEEE Conference on Computer Vision and Pattern Recognition},
  pages={6299--6308},
  year={2017}
}

@article{liu2020ms,
  title={Ms-tcn++: Multi-stage temporal convolutional network for action segmentation},
  author={Liu, Y and Cheng, MM and Li, SJ and Farha, YA and Gall, J},
  journal={IEEE Trans. Pattern Analysis and Machine Intelligence},
  pages={1--1},
  year={2020}
}

@inproceedings{du2023we,
  title={Do we really need temporal convolutions in action segmentation?},
  author={Du, Dazhao and Su, Bing and Li, Yu and Qi, Zhongang and Si, Lingyu and Shan, Ying},
  booktitle={2023 IEEE International Conference on Multimedia and Expo (ICME)},
  pages={1014--1019},
  year={2023},
  organization={IEEE}
}

@article{liu2024diffact++,
  title={Diffact++: Diffusion action segmentation},
  author={Liu, Daochang and Li, Qiyue and Dinh, Anh-Dung and Jiang, Tingting and Shah, Mubarak and Xu, Chang},
  journal={IEEE Transactions on Pattern Analysis and Machine Intelligence},
  volume={47},
  number={3},
  pages={1644--1659},
  year={2024},
  publisher={IEEE}
}

@inproceedings{brodersen2010balanced,
  title={The balanced accuracy and its posterior distribution},
  author={Brodersen, Kay Henning and Ong, Cheng Soon and Stephan, Klaas Enno and Buhmann, Joachim M},
  booktitle={2010 20th international conference on pattern recognition},
  pages={3121--3124},
  year={2010},
  organization={IEEE}
}

@inproceedings{davis2006relationship,
  title={The relationship between Precision-Recall and ROC curves},
  author={Davis, Jesse and Goadrich, Mark},
  booktitle={Proceedings of the 23rd international conference on Machine learning},
  pages={233--240},
  year={2006}
}

@inproceedings{lea2017temporal,
  title={Temporal convolutional networks for action segmentation and detection},
  author={Lea, Colin and Flynn, Michael D and Vidal, Rene and Reiter, Austin and Hager, Gregory D},
  booktitle={proceedings of the IEEE Conference on Computer Vision and Pattern Recognition},
  pages={156--165},
  year={2017}
}

\end{document}